\relax
\documentclass{article}
\usepackage{times}
\usepackage{helvet}
\usepackage{courier}
\usepackage{times}

\usepackage{algorithm}
\usepackage{algorithmic}

\usepackage{graphicx} % more modern

\usepackage{amsthm}
\usepackage{amssymb}
\usepackage{amsmath}
\usepackage{mathtools}

\usepackage{multicol}

\usepackage{graphicx} % more modern
\usepackage{caption}

\usepackage{natbib}

\usepackage{algorithm}
\usepackage{algorithmic}
\usepackage{dblfloatfix}
\usepackage{fixltx2e}

\usepackage{amsmath,amssymb,amsthm}
\usepackage[inline]{enumitem}
\usepackage{color}
\usepackage{bbm}

\usepackage[accepted]{whi2016}

\begin{document} 

\twocolumn[
	\icmltitle{Visualizing Dynamics: from t-SNE to SEMI-MDPs}
	\icmlauthor{Nir Ben Zrihem*}{bentzinir@gmail.com}
	\icmlauthor{Tom Zahavy*}{tomzahavy@campus.technion.ac.il}
%\icmladdress{Electrical Engineering
%Technion, Israel}
%\icmladdress{Electrical Engineering
%Technion, Israel}
\icmlauthor{Shie Mannor}{shie@ee.technion.ac.il}
\icmladdress{Electrical Engineering Department, The Technion - Israel Institute of Technology, Haifa 32000, Israel}
\vskip 0.3in
	]	

\begin{abstract}
Deep Reinforcement Learning (DRL) is a trending field of research, showing great promise in many challenging problems such as playing Atari, solving Go and controlling robots. While DRL agents perform well in practice we are still missing the tools to analayze their performance and visualize the temporal abstractions that they learn. In this paper, we present a novel method that automatically discovers an internal Semi Markov Decision Process (SMDP) model in the Deep Q Network's (DQN) learned representation. We suggest a novel visualization method that represents the SMDP model by a directed graph and visualize it above a t-SNE map. We show how can we interpret the agent's policy and give evidence for the hierarchical state aggregation that DQNs are learning automatically. Our algorithm is fully automatic, does not require any domain specific knowledge and is evaluated by a novel likelihood based evaluation criteria. 
\end{abstract}

\section{Introduction}
DQN is an off-policy learning algorithm that uses a Convolutional Neural Network (CNN) \cite{Krizhevsky2012} to represent the action-value function and showed superior performance on a wide range of problems \cite{mnih2015human}. The success of DQN, and that of Deep Neural Network (DNN) in general, is explained by its ability to learn good representations of the data automatically. Unfortunately, its high representation power is also making it complex to train and hampers its wide use.\\
Visualization can play an essential role in understanding DNNs. Current methods mainly focus on understanding the spatial structure of the data. 
\newpage
For example, \citet{zeiler2014visualizing} search for training examples that cause high neural activation at specific neurons, \citet{erhan2009visualizing} created training examples that maximizes the neural activity of a specific neuron and \citet{yosinski2014transferable} interpreted each layer as a group. However, none of these methods analyzed the temporal structure of the data.\\
Good temporal representation of the data can speed up the prerformence of Reinforcement Learning (RL) algorithms \cite{dietterich2000hierarchical,dean1995decomposition,parr1998flexible,hauskrecht1998hierarchical}, and indeed there is a growing interest in developing hierarchical DRL algorithms. For example, ~\citet{tessler2016deep} pre-trained skill networks using DQNs and developed a Hierarchical DRL Network (H-DRLN). Their architecture learned to control between options operating at different temporal scales and demonstrated superior performance over the vanilla DQN in solving tasks at Minecraft. \citet{kulkarni2016hierarchical} took a different approach, they manually pre-defined sub-goals for a given task and developed a hierarchical DQN (h-DQN) that is operating at different time scales. This architecture managed to learn how to solve both the sub-goals and the original task and outperformed the Vanilla DQN in the the challenging ATARI game 'Montezuma's Revenge'. Both these methods used prior knowledge about the hierarchy of a task in order to solve it. However it is still unclear how to automatically discover the hierarchy a-priori.\\
Interpretability of DQN policies is an urging issue that has many important applications. For example, it may help to distil a cumbersome model into a simple one \cite{rusu2015policy} and will increase the human confidence in the performance of DRL agents. By understanding what the agent has learned we can also decide where to grant it control and where to take over. Finally, we can improve learning algorithms by finding their weaknesses.\\
The internal model principle \citep{francis1975internal}, "Every good key must be a model of the lock it opens",  was formulated mathematically for control systems by \citet{sontag2003adaptation}, claiming that if a system is solving a control task, it must necessarily contain a subsystem which is capable of predicting the dynamics of the system. In this work we follow the same line of thought and claim that DQNs are learning an underlying spatio-temporal model of the problem, without implicitly being trained to. We identify this model as an Semi Aggregated Markov Decision Process (SAMDP), an approximation of the true MDP that allows human interpretability.\\
\citet{Zahavy2016} showed that by using hand-crafted features, they can interpret the policies learned by DQN agents using a manual inspection of a t-Distributed Stochastic Neighbor Embedding (t-SNE) map \cite{van2008visualizing}. They also revealed that DQNs are automatically learning temporal representations such as hierarchical state aggregation and temporal abstractions. On the other hand, they use a manual reasoning of a t-SNE map, a tedious process that requires careful inspection as well as an experienced eye.
However, we suggest a method that is fully automatic. Instead of manually designing features, we use clustering algorithms to reveal the underlying structure of the t-SNE map. But instead of naively applying classical methods, we designed novel time-aware clustering algorithms that take into account the temporal structure of the data. Using this approach we are able to automatically reveal the underlying dynamics and rediscover the temporal abstractions showed in \cite{Zahavy2016}. Moreover, we show that our method reveals an underlying SMDP model and confront this hypothesis qualitatively, by designing a novel visualization tool, and quantitatively, by developing likelihood criteria which we later test empirically.\\
The result is an SMDP model that gives a simple explanation on how the agent solves the task - by  decomposing it automatically into a set of sub-problems and learning a specific skill at each. Thus, we claim that we have found an internal model in DQN's representation, which can be used for automatic sub-goal detection in future work.
\section{Methodology}
\begin{enumerate}
\item \textbf{Learn :} Train a DQN agent.
\item \textbf{Evaluate :} Run the agent, record visited states, neural activations and Q-values.
\item \textbf{Reduce :} Apply t-SNE.
\item \textbf{Cluster :} Apply clustering on the data.
\item \textbf{Model :} Fit an SMDP model. Estimate the transition probabilities and reward values.
\item \textbf{Visualize :} Visualize the SMDP above the t-SNE.
\end{enumerate}
\section{From DQN to t-SNE} 
\label{DQN_tSNE}
We train DQN agents using the Vanilla DQN algorithm \cite{mnih2015human}. When training is done, we evaluate the agent at multiple episodes, using an $\epsilon$-greedy policy. We record all visited states and their neural activations, as well as the Q-values and other manually extracted features. We keep the states in their original visitation order in order to maintain temporal relations. Since the neural activations are of high order we apply t-SNE dimensionality reduction so we are able to visualize it.
\begin{figure*}[h]
\begin{center}
\includegraphics[trim=8cm 3cm 6cm 2cm,clip,width=0.6\textwidth]{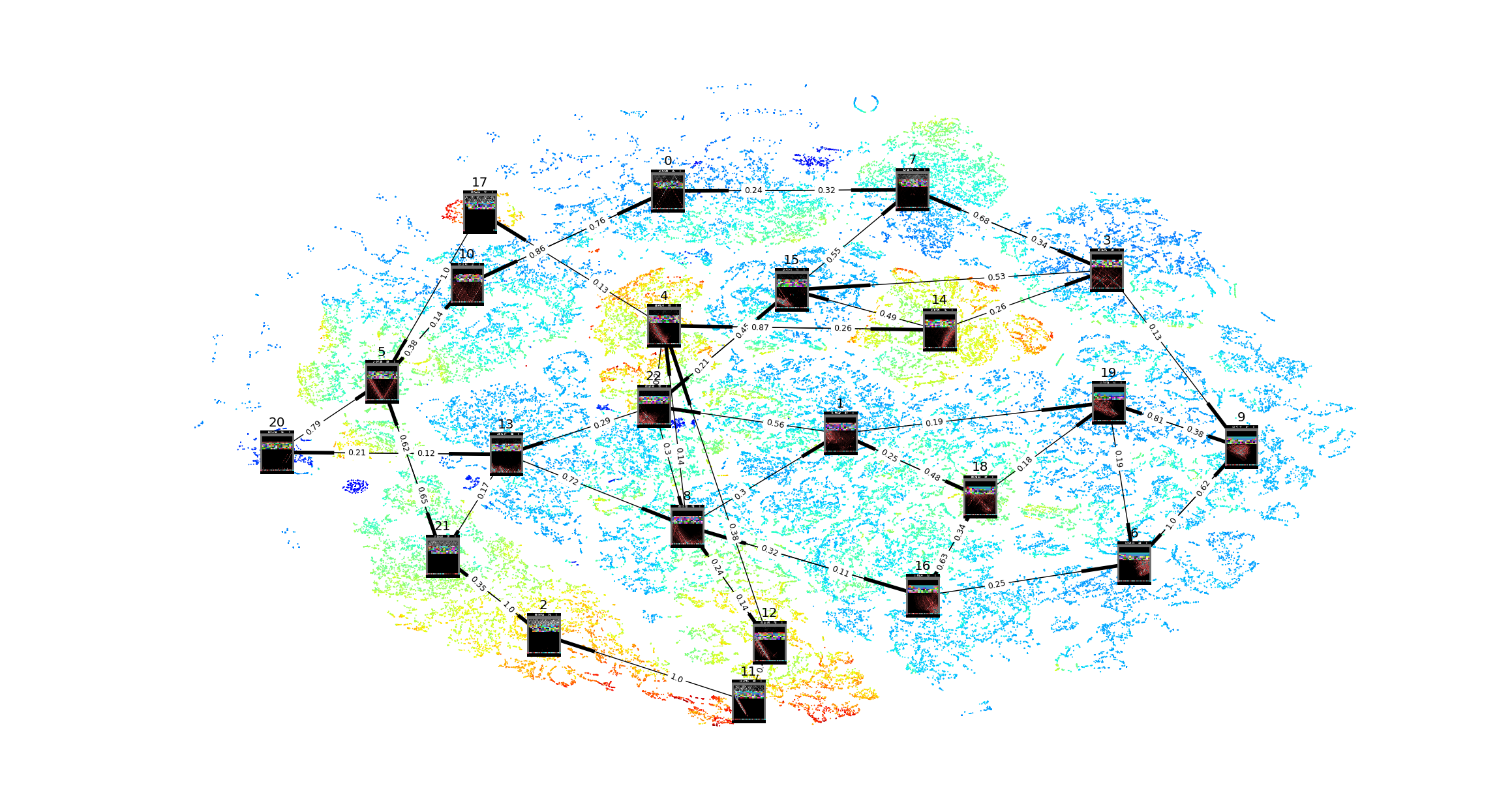} 
\caption{SMDP visualization for Breakout.}
\vspace{-0.6cm}
\label{fig:Breakout}
\end{center}
\end{figure*}

\section{From t-SNE to SMDP}
\label{tSNE_SMDP}
We define an SMDP model over the set of t-SNE points using a vector of cluster labels $C$, and a transition probability matrix $P$ where $P_{i,j}$ indicates the empirical probability of moving from cluster $i$ to $j$. We define the entropy of a model by: $e= - \sum_i \{ |C_i| \cdot \sum_j{P_{i,j} \log P_{i,j}} \}$, i.e., the average entropy over transition probability from each cluster weighted by its size.\\
We note that threw this entire paper, by an SMDP model we only refer to the induced Markov Reward Process of the DQN policy. Recall that the DQN agent is learning a deterministic policy, therefore, in deterministic environments (e.g., the Atari2600 emulator), the underlying SMDP should in fact be deterministic and an entropy minimizer.\\
The data that we collect from the DQN agent is highly correlated since it was generated from an MDP. However, standard clustering algorithms assume the data is drawn from an i.i.d distribution, and therefore result with clusters that overlook the temporal information. This results with high-entropy SMDP models that are too complicated to analayze and are not consistent with the data. For this aim, we use a variant of Kmeans that incorporates the data temporal information such that a point $x_p$ is assigned to a cluster $c_j$, if its neighbours along the trajectory are also close to $c_j$, using a temporal window.\\
We follow the analysis of \citep{hallak2013model} and define criteria to measure the fitness of a model empirically. We define the \textbf{Value Mean Square Error(VMSE)} as the normalized distance between two value estimations: $$\mbox{VMSE} = \frac{\| v^{DQN}-v^{SMDP} \|}{\|v^{DQN}\|}.$$ The SMDP value is given by 
\begin{equation}
\label{eq:smdp_value}
V_{SMDP} = ( I+\gamma^{k}P )^{-1}r
\end{equation}
and the DQN value is evaluated by averaging the DQN value estimates over all MDP states in a given cluster (SMDP state): ${v^{DQN}(c_j)}=\frac{1}{|C_j|}\sum_{i: s_i \in c_j}v^{DQN}(s_i)$ . Finally, the greedy policy with respect to the SMDP value is given by:
\begin{equation}
\label{eq:smdp_greedy_policy}
\pi_{greedy}(c_i) = \underset{j}{\mbox{argmax}} \{ R_{\sigma_{i,j}}+\gamma^{k_{\sigma_{i,j}}}v_{SMDP}(c_j) \}
\end{equation}
The \textbf{Minimum Description Length} (MDL; \citep{rissanen1978modeling}) principle is a formalization of the celebrated Occam’s Razor. It copes with the over-fitting problem for the purpose of model selection. According to this principle, the best hypothesis for a given data set is the one that leads to the best compression of the data. Here, the goal is to find a model that explains the data well, but is also simple in terms of the number of parameters. In our work we follow a similar logic and look for a model that best fits the data but is still “simple”.\\
Instead of considering "simple" in terms of the number of parameters, we measure the simplicity of the spatio-temporal state aggregation. For spatial simplicity we define the Inertia: $I = \sum_{i=0}^{n}\min_{\mu_j \in C}(||x_j - \mu_i||^2)$ which measures the variance of MDP states inside a cluster (AMDP state). For temporal simplicity we define the entropy: $e= - \sum_i \{ |C_i| \cdot \sum_j{P_{i,j} \log P_{i,j}} \}$ , and the \textit{Intensity Factor} which measures the fraction of in/out cluster transitions: $F = \sum_j \frac{P_{jj}}{\sum_i P_{ji}}.$
\section{Visualization: fusing SMDP with t-SNE}
\label{Visualization}
In Section~\ref{DQN_tSNE} we explained how to create a t-SNE map from DQN's neural activations and in Section~\ref{tSNE_SMDP} we showed how to automatically design an SMDP model using temporal-ware clustering methods. In this section we explain how to fuse the SMDP model with the t-SNE map for a clear visualization of the dynamics.\\
In our approach, an SMDP is represented by a directed graph. Each SMDP state is represented by a node in the graph and corresponds to a cluster of t-SNE points (game states). In addition, the transition probabilities between the SMDP states are represented by weighted edges between the graph nodes. We draw the graph on top of the t-SNE map such that it reveals the underlying dynamics. Choosing a good layout mechanism to represent a graph is a hard task when dealing with high dimensional data \cite{tang2016visualizing}. We consider different layout algorithms for the position of the nodes, such as the spring layout that position nodes by using the Fruchterman-Reingold force-directed algorithm and the spectral layout that uses the eigenvectors of the graph Laplacian \cite{hagberg-2008-exploring}. However, we found out that simply positioning each node at the average coordinates of each t-SNE cluster gives a more clear visualization. The intuition behind it is that the t-SNE algorithm was planned to solve the crowding problem and therefore outputs clusters that are well separated from each other.
\section{Experiments}
\label{Experiments}
\begin{figure*}
\begin{center}
\includegraphics[trim=2cm 1cm 2cm 1cm,clip,width=0.6\textwidth]{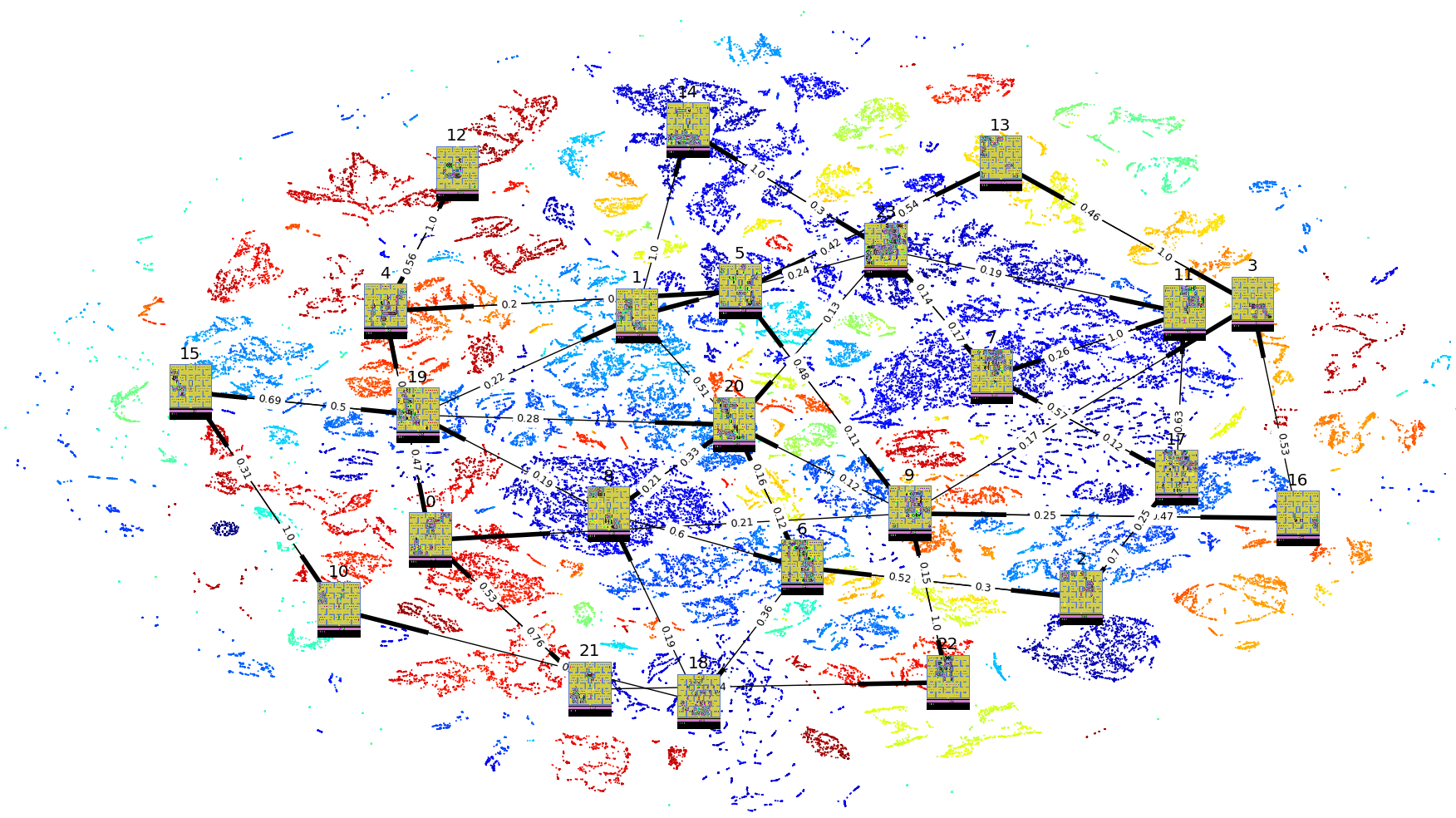} 
\caption{SMDP visualization for Pacman.}
\vspace{-0.6cm}
\label{fig:Pacman}
\end{center}
\end{figure*} 
\textbf{Experimental set-up}.
We evaluated our method on two Atari2600 games, Breakout and Pacman. For each game we collected 120k game states. We apply the t-SNE algorithm directly on the collected neural activations of the last hidden layer, similar to \citet{mnih2015human}. The input $X\in\mathbf{R}^{120k\times512}$ consists of $120k$ game states with $512$ features each (the size of our DQN last layer). Since this data is relatively large, we pre-processed it using Principal Component Analysis to dimensionality of 50 and used the Barnes Hut t-SNE approximation  \cite{van2014accelerating}. The input $X\in\mathbf{R}^{120k\times3}$ to the clustering algorithm consists of $120k$ game states with $3$ features each (two t-SNE coordinates and the Value estimate). We applied the Spatio-Temporal Cluster Assignment with  \textit{k=20} clusters and \textit{w=2} temporal window size. We run the algorithm for $160$ iterations and choose the best SMDP in terms of minimum entropy (we will consider other measures in future work). Finally we visualize the SMDP using the visualization method explained in Section~\ref{Visualization}.\\
\textbf{Simplicity}.
Looking at the resulted SMDPs it is interesting to note that the transition probability matrix is very sparse, i.e., the transition probability from each state is not zero only for a small subset of the states, thus, indicating that our cluster are located in time. Inspecting the mean image of each cluster we can see that the clusters are also highly spatially located, meaning that the states in each cluster share similar game position. Figure~\ref{fig:Breakout} shows the SMDP for \textbf{Breakout}. The mean image of each cluster shows us the ball location and direction (in red), thus characterizes the game situation in each cluster. We also observe that states with low entropy follow a well defined skill policy. For example cluster 10 has one main transition ans show a well defined skill of carving the left tunnel (see the mean image). In contrast, clusters 6 and 16 has transitions to more clusters (and therefore higher entropy) and a much less defined skill policy (presented by its relatively confusing mean state). Figure~\ref{fig:Pacman} shows the SMDP for \textbf{Pacman}. The mean image of each cluster shows us the agent's location (in blue), thus characterizes the game situation in each cluster. We can see that the agent is spending its time in a very defined areas in the state space at each cluster. For example, cluster 19 it is located in the north-west part of the screen and in cluster 9 it is located in south-east. We also observe that clusters with more transitions, e.g., clusters 0 and 2, suffer from less defined mean state.\\
\begin{figure}
\centering
\includegraphics[trim=2cm 0cm 0cm 1cm,clip,width=0.45\textwidth]{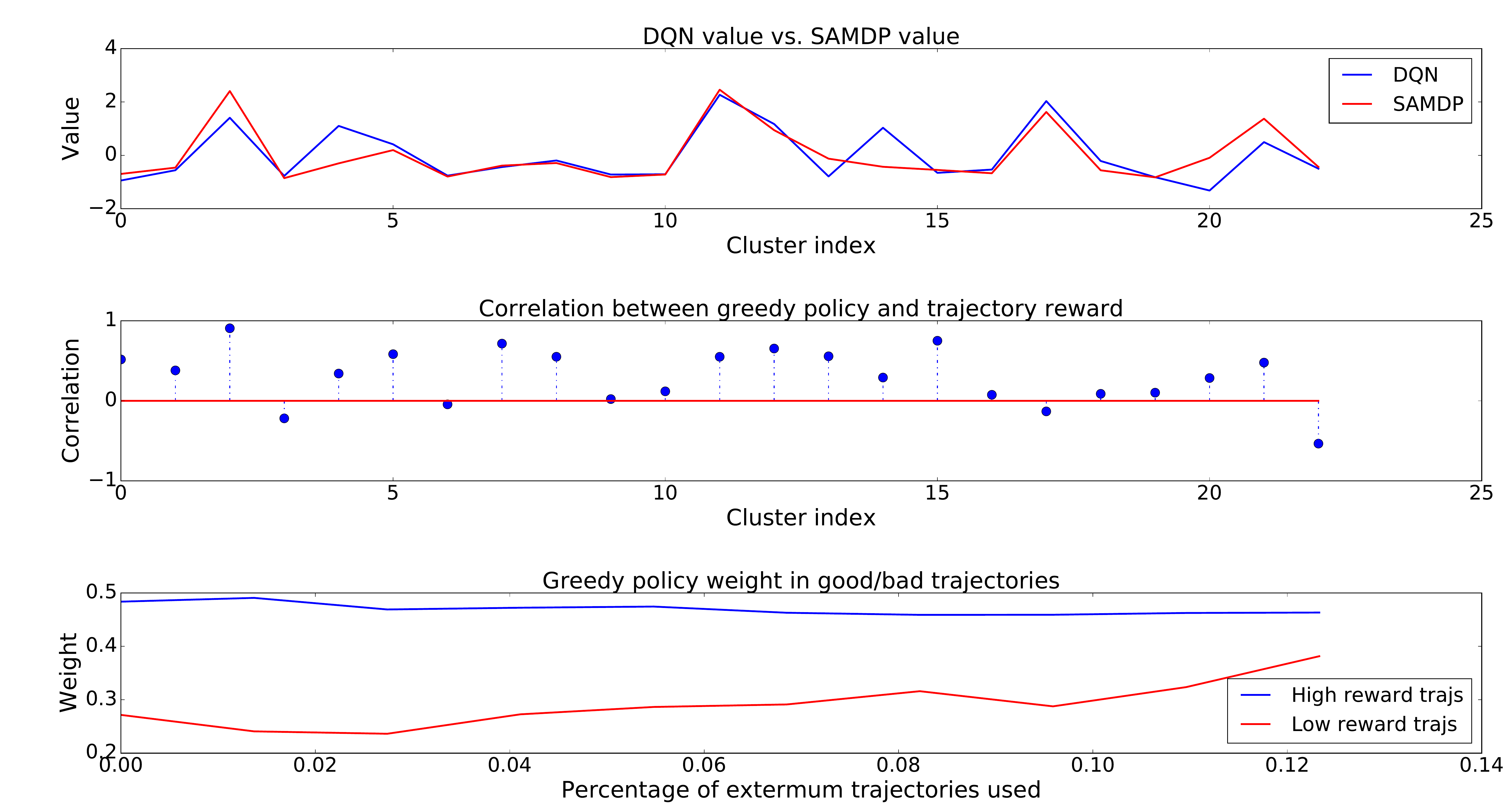}
\vspace{-0.6cm}
\caption{\textbf{Model Evaluation.} \textbf{Top:} Value function consistency. \textbf{Center:} greedy policy correlation with trajectory reward. \textbf{Bottom:} top (blue), least (red) rewarded trajectories.}
\vspace{-0.6cm}
\label{fig:model_consis}
\end{figure}
\textbf{Model Evaluation.} We evaluate our model using three different methods. First, the VMSE criteria (Figure~\ref{fig:model_consis}, top): high correlation between the DQN values and the SMDP values gives a clear indication to the fitness of the model to the data. Second, we evaluate the correlation between the transitions induced by the policy improvement step and the trajectory reward $R^j$. To do so, we measure $P_i^j:$ the empirical distribution of choosing the greedy policy at state $c_i$ in that trajectory. Finally we present the correlation coefficients at each state: $corr_i = corr(P_i^j,R^j)$ (Figure~\ref{fig:model_consis}, center). Positive correlation indicates that following the greedy policy leads to high reward. Indeed for most of the states we observe positive correlation, supporting the consistency of the model. The third evaluation is close in spirit to the second one. We create two transition matrices $T^+,T^-$ using k top-rewarded trajectories and k least-rewarded trajectories respectively. We measure the correlation of the greedy policy $T^G$ with each of the transition matrices for different values of k (Figure~\ref{fig:model_consis} bottom). As clearly seen, the correlation of the greedy policy and the top trajectories is higher than the correlation with the bad trajectories.
\section{Discussion}
\label{Discussion}
In this work we considered the problem of visualizing dynamics. Starting with a t-SNE map of the neural activations of a DQN and ending up with an SMDP model describing the underlying dynamics. We developed clustering algorithms that take into account the temporal aspects of the data and defined quantitative criteria to rank candidate SMDP models based on the likelihood of the data and an entropy simplicity term. Finally we showed in the experiments section that our method can successfully be applied on two Atari2600 benchmarks, resulting in a clear interpretation for the agent policy.\\
Our method is fully automatic and does nor require any manual or game specific work. We note that this is a work in progress, it is mainly missing the quantitative results for the different likelihood criteria. In future work we will finish to implement the different criteria followed by the relevant simulations.
\newpage
\small
\bibliography{paper_bib}

\bibliographystyle{icml2016}
\clearpage

\normalsize
\end{document}